\documentclass[11pt]{arxform}
\usepackage{times}

\usepackage{soul}
\usepackage{url}
\usepackage[hidelinks]{hyperref}
\usepackage[utf8]{inputenc}
\usepackage[small]{caption}
\usepackage{graphicx}
\usepackage{booktabs}
\usepackage{latexsym}

\usepackage{amsthm}
\usepackage{mathtools}
\usepackage{stmaryrd}
\usepackage{amsfonts} 
\usepackage{nicefrac}
\usepackage{booktabs} 
\usepackage{multirow} 
\usepackage{multicol} 
\usepackage{algorithm,algorithmic}
\usepackage{comment}
\usepackage[disable]{todonotes}
\usepackage{bm}
\usepackage{dsfont,tikz,ifthen,subfigure}
\usepackage{tikz}
\usepackage{pgfplots}	

\pgfplotsset{compat=newest}
\pgfplotsset{/pgfplots/error bars/error bar style={very thick}}
\pgfplotsset{
  every axis plot/.append style={very thick, black},
}

\newcommand{\Prob}{\vec{p}}

\newcommand{\given}{\, | \,}

\renewcommand{\vec}[1]{\boldsymbol{#1}}
\newcommand*{\defeq}{\mathrel{\vcenter{\baselineskip0.5ex \lineskiplimit0pt
			\hbox{\footnotesize.}\hbox{\footnotesize.}}}%
	=}
\newcommand{\defi}{\defeq}

\newcommand{\fromto}{\longrightarrow}

\newcommand{\cX}{\mathcal{X}}
\newcommand{\cL}{\mathcal{L}}
\newcommand{\cY}{\mathcal{Y}}

\newcommand{\cD}{\mathcal{D}}
\newcommand{\bx}{\boldsymbol{x}}
\newcommand{\by}{\boldsymbol{y}}

\newcommand{\bY}{\mathbf{Y}}
\newcommand{\bh}{\boldsymbol{h}}

\newcommand{\loss}{\ell}

\begin{document}

\title{On Aggregation in Ensembles of Multilabel Classifiers}
\titlerunning{On Aggregation in Ensembles of Multilabel Classifiers}

	\author{Eyke H{\"u}llermeier}

\author{ Vu-Linh Nguyen\inst{1} \and Eyke H\"ullermeier\inst{1} \and Michael Rapp\inst{2}  \and \\ Eneldo Loza Menc\'ia\inst{2} \and Johannes F\"urnkranz\inst{3}}
\authorrunning{V.-L. Nguyen et al.}

\institute{
 Heinz Nixdorf Institute and Department of Computer Science Paderborn University, Germany
\email{vu.linh.nguyen@uni-paderborn.de}, \email{eyke@upb.de}
\and
Knowledge Engineering Group, TU Darmstadt, Germany
\email{mrapp@ke.tu-darmstadt.de}, \email{eneldo@ke.tu-darmstadt.de}
\and
Computational Data Analytics Group, JKU Linz, Austria
\email{juffi@faw.jku.at}
}

	\maketitle

	\begin{abstract}
 While a variety of ensemble methods for multilabel classification have been proposed in the literature, the question of how to aggregate the predictions of the individual members of the ensemble has received little attention so far. In this paper, we introduce a formal framework of ensemble multilabel classification, in which we distinguish two principal approaches: ``predict then combine" (PTC), where the ensemble members first make loss minimizing predictions which are subsequently combined, and ``combine then predict" (CTP), which first aggregates information such as marginal label probabilities from the individual ensemble members, and then derives a prediction from this aggregation. While both approaches generalize voting techniques commonly used for multilabel ensembles, they allow to explicitly take the target performance measure into account. Therefore, concrete instantiations of CTP and PTC can be tailored to concrete loss functions. Experimentally, we show that standard voting techniques are indeed outperformed by suitable instantiations of CTP and PTC, and provide some evidence that CTP performs well for decomposable loss functions, whereas PTC is the better choice for non-decomposable losses. 

	\end{abstract}


\section{Introduction}

The setting of \emph{multilabel classification} (MLC), which generalizes standard multi-class classification by relaxing the assumption of mutual exclusiveness of classes, has received a lot of attention in the recent machine learning literature---we refer to  \cite{tsoumakas:2010} and \cite{zhang:2014} for comprehensive survey articles on this topic. 

Like for other types of classification problems, the idea of \emph{ensemble learning}  \cite{dietterich:2000}
has also been applied to MLC (cf.\ Section~\ref{sec:EMLC}). However, somewhat surprisingly, the question of how to aggregate the predictions of the individual members of an ensemble has so far received little attention in MLC. Instead, most approaches are based on simple voting techniques, which are typically applied in a label-wise manner: For each label, the predictions---either binary predictions of relevance or, more generally, label probabilities---of all ensemble members are collected, averaged, and thresholded to obtain a final prediction for this label.

An obvious disadvantage of this simple approach is that the aggregation is independent of the underlying performance measure, i.e., the aggregation procedure is not tailored to a specific loss function. This, however, would supposedly be important: In contrast to standard classification, where a loss function compares a predicted class label with a ground truth, an MLC loss compares a \emph{subset} of labels predicted to be relevant with a ground-truth subset. As there are various ways in which subsets can be compared with each other, a wide spectrum of loss functions is commonly used in MLC, and it is well known that different losses may call for different (Bayes-optimal) predictions \cite{dembczynski:2012,dembczynski:2012:PCC}. Naturally, the idea of customizing an MLC predictor to a specific loss function should not only be considered at the level of individual predictors, but also at the level of the ensemble as a whole, and hence also concern the way in which the predictions are combined.

In this paper, we study the problem of aggregation in ensembles of multilabel classifiers (EMLC) in a systematic way. To this end, we introduce a formal framework, in which we distinguish two principal approaches: ``predict then combine" (PTC), where the ensemble members first make loss minimizing predictions which are then combined, and ``combine then predict" (CTP), which first aggregates information such as marginal label probabilities from the individual ensemble members, and then derives a prediction from this aggregation. While both approaches generalize common voting techniques as mentioned above, they also include more general variants and, moreover, allow one to explicitly take the target loss into account. In other words, concrete instantiations of CTP and PTC can be tailored to concrete loss functions. In an extensive experimental study, we demonstrate that such loss-based aggregation functions do indeed outperform simple voting techniques, and also investigate the question which type of aggregation is more suitable for which loss functions.


\section{Multilabel Classification}

Let $\cX$ denote an instance space, and let $\cL= \{\lambda_1, \ldots, \lambda_K\}$ be a finite set of class labels. We assume that an instance $\bx \in \cX$ is (probabilistically) associated with a subset of labels $\Lambda = \Lambda(\bx) \in 2^\cL$; this subset is often called the set of relevant labels, while the complement $\cL \setminus \Lambda$ is considered as irrelevant for $\bx$. We identify a set $\Lambda$ of relevant labels with a binary vector $\by = (y_1, \ldots, y_K)$, where $y_k = \llbracket \lambda_k \in \Lambda \rrbracket$.\footnote{$\llbracket \cdot \rrbracket$ is the indicator function, i.e., $\llbracket A \rrbracket = 1$ if the predicate $A$ is true and $=0$ otherwise.} By $\cY = \{0,1\}^K$ we denote the set of possible labelings.  

We assume observations to be realizations of random variables generated independently and identically (i.i.d.) according to a probability distribution $\Prob$ on $\cX \times \cY$, i.e., an observation $\by=(y_1,\ldots, y_K)$ is the realization of a corresponding random vector $\bY = (Y_1, \ldots, Y_K)$. We denote by $\Prob(\mathbf{Y} \given \bx)$ the conditional distribution of $\bY$ given $\mathbf{X}=\bx$, and by $\Prob_k(Y_k \given \bx)$ the corresponding marginal distribution of $Y_k$:
\begin{equation}\label{eq:marginal}
\Prob_k( b \given \bx) = 
\sum_{\by\in\cY: y_k = b} \Prob(\by \given \bx) \, .
\end{equation}
Moreover, we denote by $p_k = p_k(\bx) = \Prob_k( 1 \given \bx)$ the probability of relevance of the label $\lambda_k$. 

Given training data in the form of a finite set of observations
\begin{equation}\label{eq:trainingdata}
\cD = \big\{ (\bx_n,\by_n) \big\}_{n=1}^N  \subset \cX \times \cY \, ,
\end{equation}
drawn independently from $\Prob(\mathbf{X},\mathbf{Y})$, the goal in MLC is to learn a predictive model in the form of a multilabel classifier $\bh$, which is a mapping $\cX \fromto \cY$ that assigns a (predicted) label subset to each instance $\bx\in \cX$. Thus, the output of a classifier $\bh$ is a vector of predictions
\begin{equation}\label{eq:h}
\bh(\bx) = (h_1(\bx), \ldots , h_K(\bx)) \in \{ 0,1 \}^K \,  ,
\end{equation}
also denoted as $\hat{\by} = (\hat{y}_1, \ldots , \hat{y}_K)$.

\subsection{MLC Loss Functions}

The main goal in MLC is to induce predictions (\ref{eq:h}) that generalize well beyond the training data (\ref{eq:trainingdata}), i.e., predictions 
\begin{equation}\label{eq:lossmini}
\hat{\by} =  \operatorname*{argmin}_{\bar{\by}}
\sum_{\by \in \cY} \ell(\by, \bar{\by}) \, \Prob(\by \given \bx) \, ,
\end{equation}
that minimize the expected loss with respect to a specific MLC loss function $\ell:\, \cY^2 \longrightarrow \mathbb{R}$. Two important loss functions, both generalizing the standard 0/1 loss commonly used in classification, are the \emph{Hamming loss} and the \emph{subset 0/1~loss}:
\begin{align}
\label{eq:hamming}
\loss_H(\by, \hat{\by}) & \defeq \frac{1}{K}
\sum_{k=1}^K  \, \llbracket y_k \neq  \hat{y}_k \rrbracket \enspace , \\
\loss_S(\by, \hat{\by}) & \defeq \llbracket \by \neq  \hat{\by} \rrbracket \, . \label{eq:subset}
\end{align}

The (\emph{instance-wise}) \emph{F-measure} compares a set of predicted labels to a corresponding set of ground-truth labels via the harmonic mean of precision and recall:
\begin{equation}
\label{eq:Fmeasure}
F(\by, \hat{\by})  = \frac{2 \sum_{k=1}^K \hat{y}_k \, y_k}{\sum_{k=1}^K \hat{y}_k + \sum_{k=1}^K y_k} \, .
\end{equation}

The goal of classification algorithms in general is to capture dependencies between input features and the target variable. In MLC, dependencies may not only exist between the features and each target, but also between the targets $Y_1, \ldots , Y_K$ themselves. The idea to improve predictive accuracy by capturing such dependencies is a driving force in research on multilabel classification. 

Not all loss functions capture label dependencies to the same extent:
A \emph{decomposable loss} can be reduced to loss functions for the individual labels, i.e., it can be expressed in the form 
\begin{equation}\label{eq:decoml}
\loss(\by , \hat{\by} ) = \sum_{k=1}^K \loss_k(y_k , \hat{y}_k ) \, ,
\end{equation}
with suitable binary loss functions $\loss_k:\, \{0,1\}^2 \fromto \mathbb{R}$. A \emph{non-decomposable loss} does not permit such a representation. It can be shown that, for making optimal predictions $\hat{\by} = \bh(\bx)$ which minimize the expected loss, knowledge about the marginals (\ref{eq:marginal}) is sufficient in the case of a decomposable loss (such as Hamming), but not in the case of a non-decomposable loss  \cite{dembczynski:2012}. Instead, if a loss is non-decomposable, higher-order probabilities are needed, and in the extreme case even the entire distribution $\Prob(\mathbf{Y} \given \bx)$ (like in the case of the subset 0/1 loss). 

On an algorithmic level, this means that MLC with a decomposable loss can be tackled by what is commonly called binary relevance (BR) learning, i.e., by learning one binary classifier for each individual label, whereas non-decomposable losses call for more sophisticated learning methods that are able to take label dependencies into account.

\subsection{Risk Minimization}
\label{sec:riskm}

In the most general case, the problem of finding a risk-minimizing (Bayes-optimal) prediction  is tackled by producing a prediction $\Prob(\cdot \given \bx)$ of the conditional joint distribution of labelings, and explicitly solving (\ref{eq:lossmini}) as a combinatorial optimization problem. Obviously, this approach is infeasible unless the number of class labels is very low. 
Fortunately, the problem can be solved more efficiently for specific loss functions, including those considered in this paper. 

In the case of the \emph{Hamming loss}, the Bayes-optimal prediction can be obtained by thresholding the marginal probabilities, regardless of whether the labels are independent or not:
\begin{equation}\label{eq:bayes-optimal-hamming}
\hat{y}_k = \left\{ \begin{array}{cl}
0 & \text{if } p_k(\vec{x}) \leq \nicefrac{1}{2}\\
1 & \text{if } p_k(\vec{x}) > \nicefrac{1}{2}
\end{array}\right.
\end{equation}
Thus, it is sufficient to have good estimates for the marginal probabilities, which can be accomplished by simple techniques such as binary relevance \cite{dembczynski:2012}.

For \emph{subset 0/1 loss}, the Bayes-optimal prediction is not the marginal but the \emph{joint} mode of the distribution $\Prob(\cdot \given \bx)$: 
\begin{equation*}
\hat{\by} \in \operatorname*{argmax}_{\bar{\by}\in \cY} \Prob(\bar{\by} \given \bx) \, .
\end{equation*}
In this case, label dependence needs to be taken into account for optimal performance. 

The \emph{F-measure} is in a sense in-between these two extremes. It can be shown that, while the entire distribution $\Prob(\cdot \given \bx)$ is not needed to find a Bayes-optimal prediction for this measure, marginal probabilities (\ref{eq:marginal}) do not suffice either. Instead, probabilities on pairwise label combinations are required in the general case, whereas under the assumption of conditional label independence, marginal probabilities again provide sufficient information \cite{waegeman:2014}.

\section{Ensembles of MLC}
\label{sec:EMLC}

In general, an ensemble approach to multilabel classification (EMLC) learns a set of $M$ multilabel classifiers, each of which predicts a binary label vector $\hat{\by}_j$. Given a query instance $\bx \in \cX$, these are then aggregated into a final prediction $\hat{\by} =  \operatorname{agg}(\hat{\by}_1, \dots, \hat{\by}_M)$. For this aggregation, variants of label-wise \emph{majority voting (MV)} are typically used:
\begin{itemize}
\item[•] \textbf{Binary majority voting (BMV)} assigns to each label $\lambda_k \in \mathcal{L}$ the prediction given by the majority of the classifiers:
\begin{equation}\label{eq:BMv}
\hat{y}_k \defi  \operatorname*{argmax}_{y_k \in \{0,1\}} \sum_{j=1}^M  \llbracket y_k =  \hat{y}_{j,k} \rrbracket \, .    
\end{equation}
\item[•] \textbf{Graded majority voting (GMV)}, also known as \emph{weighted voting}, adds up 
confidence scores $\Prob_j = (p_{j,1}, p_{j,2}, \ldots, p_{j,K})$ for each label $\lambda_k \in \mathcal{L}$:
\begin{equation}\label{eq:GMV}
\hat{y}_k \defi  \operatorname*{argmax}_{y_k \in \{0,1\}} \sum_{j=1}^M p_{j,k}^{y_k} (1 - p_{j,k})^{1- y_k} \, .    
\end{equation}
\end{itemize}

Several ensemble-based multi-label classifiers have been tried in the literature, which typically use the above-mentioned voting techniques for combining the predictions of the ensemble members \cite{gharroudi:2017,gharroudi:2015,madjarov:2012,shi:2011,tsoumakas:2007}.  While we aim at optimizing the predictions for a particular loss function, a
different line of work---orthogonal to our approach---aims at simultaneously optimizing for multiple loss functions \cite{MLC-MOO2,MLC-MOO1}. In the following, we briefly recall some commonly used EMLC methods, which will serve as baselines in our experimental evaluation.  We refer to  \cite{moyano:2018} for an extensive discussion on ensembles of MLC classifiers.

\begin{itemize}
\item[•] \textbf{Ensembles of Binary Relevance Classifiers (EBR)} use bagging  \cite{breiman:1996} to construct $K$ independent ensembles of binary classifiers, one for each label $\lambda_k \in \mathcal{L}$ \cite{tsoumakas:2009}.
At prediction time, the predictions of these classifiers are combined for each label using majority voting, as is commonly used in bagged ensembles. Obviously, like all BR methods, EBR ignores any relationships between the labels and implicitly assumes them to be independent. Moreover, EBR is computationally expensive, since $K\cdot M$ classifiers are required in order to have an ``actual ensemble'' of cardinality~$M$. 

\item[•] \textbf{Ensembles of Classifier Chains (ECC).} The classifier chains (CC) method  \cite{read:2009} also trains $K$ binary classifiers $h_k$, $k \in [K] \defeq \{1, \ldots , K\}$, one for each label. Yet, to capture label dependencies, $h_k$ is trained on an augmented input space $\mathcal{X} \times \{0,1\}^{k-1}$, taking the (binary) values of the $k-1$ previous labels as additional attributes. More specifically, $h_k$  predicts $\hat{y}_{\sigma(k)} \in \{0,1\}$~using 
\begin{align*}
\left(\bx ,\hat{y}_{\sigma(1)}, \hat{y}_{\sigma(2)}, \ldots, \hat{y}_{\sigma(k-1)}\right) \in \mathcal{X} \times \{0,1\}^{k-1} 
\end{align*}
as input, where $\sigma$ is some permutation of $[K]$. 

Practically, it turns out that the order of labels on the chain, defined by $\sigma$, has an impact on predictive performance  \cite{cheng:2010,read:2011}. As finding an optimal order appears to be difficult, \cite{read:2011} suggest to use an ensemble of CCs over a (randomly chosen) set of permutations and combine their predictions. In the original CC, the final prediction is derived in a label-wise manner using BMV. In a probabilistic variant of CC, we allow each classifier $h_k$, $k \in [K]$, to produce a score in $[0,1]$, namely an estimation of the conditional probability
\begin{equation}\label{eq:PCCprobabilistic}
\Prob\left(y_{\sigma(k)} = 1 \, | \, \bx , y_{\sigma(1)}, \ldots , y_{{\sigma(k-1)}}\right) \, .  
\end{equation}
The score \eqref{eq:PCCprobabilistic} can be seen as a ``dependent'' marginal probability, i.e., a marginal probability which to some extent takes label dependence into account.  

\item[•] \textbf{Ensembles of Multi-Objective Decision Trees (EMODT)} are a computationally efficient EMLC method \cite{kocev:2007}. Similar to conventional decision trees (DT) \cite{DecisionTrees-Survey,quinlan:1986}, a multi-objective decision tree (MODT) partitions the instance space $\mathcal{X}$ into (axis-parallel) regions $R_1, \ldots , R_L$ (i.e., $\bigcup_{i=1}^L R_i = \mathcal{X}$ and $R_i \cap R_j = \emptyset$ for $i \neq j$), corresponding to individual leaves of the tree. In a probabilistic setting, each leaf of the MODT is associated with a complete marginal probability vector, where the marginal probability corresponding to a particular label is simply estimated as the proportion of the training instances in the leaf for which the label is relevant.  The binary label vector predicted by an EMODT can be derived with GMV on the probability vectors provided by the individual MODTs in a label-wise manner. Due to the label-wise voting, EMODT is also tailored to decomposable performance measures.
\end{itemize}

\section{A Formal Framework}
\label{sec:framework}

In the following, we define a formal framework for ensembles of multi-label classifiers.

\subsection{Intermediate Relevance Information}

Most MLC methods are two-step approaches in the sense that, prior to making a final prediction $\hat{\by} \in \cY$, 
intermediate results about the relevance of labels, their interdependencies, or similar information is compiled. We refer to such results as \emph{relevance information}, which we distinguish from the final prediction. Important examples include the~following:
\begin{itemize}
\item[•] Estimates of \emph{marginal probabilities}  (\ref{eq:marginal}), which provide important information for the minimization of decomposable loss functions, or loss minimization in the case of label independence.
\item[•] The entire \emph{joint distribution} $\Prob( \cdot \given \bx)$, which might be needed for the minimization of non-decomposable losses in cases where the labels are not independent.
\item[•] \emph{Probability estimates of a more general kind}. For example, \cite{waegeman:2014} require the probabilities $\Prob( y_k = 1, s_{\vec{y}} = s)\, , k, s \in [K]$, for loss minimization in the case of the F-measure.
\end{itemize}
In general, of course, the relevance information does not need to be probabilistic, but might be of a more general nature.  

\subsection{CTP versus PTC}

\begin{figure}[t!]
	\begin{center}
		\includegraphics[width=\columnwidth]{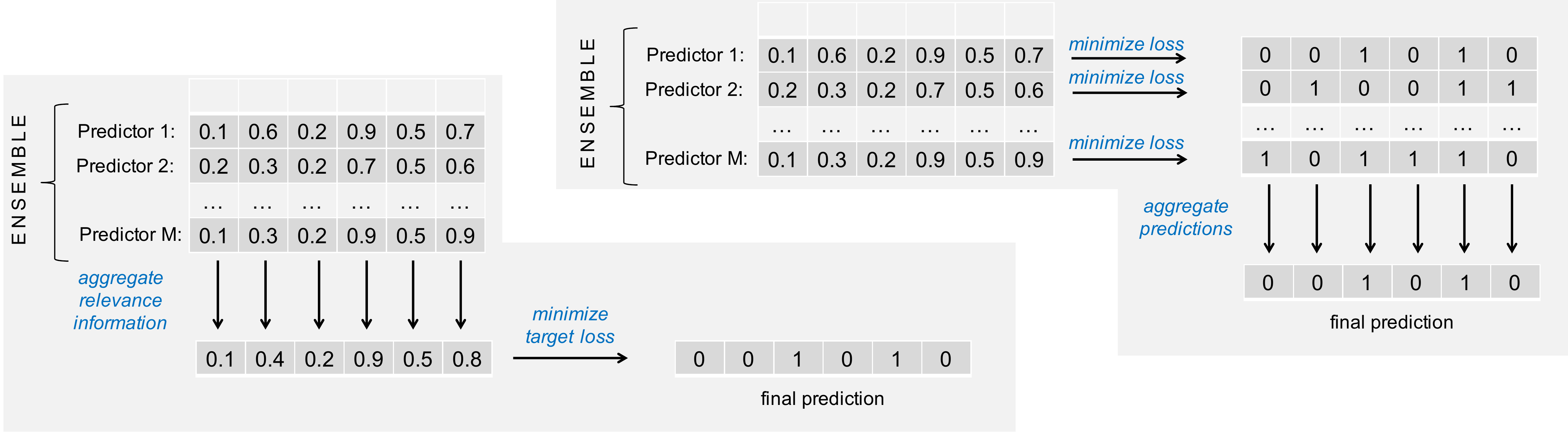} 		
		\caption{Illustration of the ``combine then predict'' (red) and ``predict then combine'' (blue) approaches for the case where relevance information consists of marginal~probabilities.}
		\label{fig:ctp}
		
	\end{center}
\end{figure}

In the context of ensemble learning, an important distinction between methods can be made depending on whether the relevance information provided by the different ensemble members is combined first, and a prediction is obtained afterwards, or whether individual predictions are produced first and then combined into an overall prediction (see Fig.\ \ref{fig:ctp} for an illustration). We refer to the former as ``combine then predict'' (CTP) and the latter as ``predict then combine'' (PTC).

In CTP, the relevance information $\mathcal{R} = \{ R_1, \ldots , R_M \}$ provided by the individual ensemble members is first combined into a single condensed representation 
\begin{equation}
R = \operatorname{CTP-agg} \big( R_1, \ldots , R_M \big) \, .
\end{equation}
Then, a final prediction $\hat{\by}$ is produced on the basis of this representation, typically (though not necessarily) taking the underlying target loss $\ell$ into account, i.e., minimizing expected loss with regard to $\ell$ (cf.\ Section \ref{sec:riskm}). Denoting the prediction step by $\operatorname{Pred}_\ell$, this can be written compactly as follows: 
\begin{equation}
\hat{\by} = \operatorname{Pred}_\ell \Big( \operatorname{CTP-agg} \big( R_1, \ldots , R_M \big) \Big) \,.
\end{equation}

In PTC, each member of the ensemble first predicts a (loss minimizing) label combination  $\hat{\by}_j = \operatorname{Pred}_\ell(R_j)$. Then, in a second step, these predictions $\hat{\by}_m$, $m \in [M]$, are combined into an overall prediction $\hat{\by}$ using a suitable aggregation function:
\begin{equation}\label{eq:ptcpred}
\hat{\by} = \operatorname{PTC-agg} \Big( \operatorname{Pred}_\ell(R_1), \ldots , 
\operatorname{Pred}_\ell(R_M) \Big) \nonumber \,.
\end{equation}

Note that the commonly used techniques of weighted and binary voting as described in Section~\ref{sec:EMLC} can be seen as specific instantiations of CTP and PTC: Binary majority voting (BMV) first maps vectors of marginal label probabilities into label predictions, which are then combined via majority voting, and is thus an instance of PTC. Graded majority voting (GMV) first adds up the label probabilities into a single vector of marginal label probabilities, which are then thresholded for a final prediction, and is thus a special case of CTP. However, both voting methods are oblivious to specific loss~functions. 

\subsection{Aggregation in CTP}

The information that needs to be combined in both approaches, CTP and PTC, is of different nature. Thus, one may expect different types of aggregation functions to be suitable. In particular, relevance information to be combined in CTP is often \emph{gradual} and represented in numerical form\,---\,probability estimates is again a typical example. Information of that kind is often reasonably combined through \emph{averaging}. For instance, the arithmetic mean 
\begin{equation}\label{eq:aggmean}
\bar{p}_i = \frac{1}{M} \sum_{j=1}^M p_{i,j} \, ,
\end{equation} 
produced by the ensemble members for the label $\lambda_i$, will be an improved estimate of the true marginal probability of that label. Of course, aggregation functions other than the arithmetic mean are also conceivable; for example, the median is known to be more robust toward outliers. 

Moreover, aggregation does not necessarily need to be label-wise as in (\ref{eq:aggmean}). Instead, it depends on what kind of relevance information is produced in the first place. Imagine, for example, that each ensemble member yields an estimate $\hat{\Prob}_j(\cdot \given \bx)$ of the joint label distribution on $\cY$. Aggregation should then be done at the same level, and averaging is again an obvious way for doing so: 
\begin{align*}
\hat{\Prob}(\by \given \bx) = \frac{1}{M}\sum_{j=1}^M \hat{\Prob}_j(\by \given \bx) \, , \forall \by \in \cY \, .    
\end{align*}
As already said, an approach of that kind might be advantageous in the case of non-decomposable losses, although it will not be tractable in general.

\subsection{Aggregation in PTC}

In PTC, the problem is to combine (binary) predictions. 
More specifically, recalling the goal to minimize a given target loss $\ell$, the problem can be stated as follows: Given predictions $\hat{\by}_1 , \ldots , \hat{\by}_M$, which are all supposed to minimize $\ell$ in expectation, what is a Bayes-optimal overall prediction $\hat{\by}$? The answer to this question is far from obvious and, to the best of our knowledge, has not been studied systematically in the literature so far. In fact, a formal analysis of this problem  probably presupposes additional assumptions about how the predictions (\ref{eq:ptcpred}) may differ from the true Bayes-optimal prediction (obviously, they cannot all be Bayes-optimal at the same time, unless they all coincide).

In any case, it should be clear that averaging will be less suitable. First of all, binary predictions $\hat{\by}$ are \emph{discrete} entities, and by averaging them one does not again end up with a discrete entity. This is to some extent comparable to the difference between ensemble \emph{regression} (numerical case) and ensemble \emph{classification} (categorical case): While arithmetic averaging is often used in the former, counting or ``voting'' techniques are more commonly applied in the latter. Second, even when solving this technical issue by turning an average into a discrete entity, for example by thresholding, undesirable effects might be produced, as shown by a simple example,
in which the (conditional) ground-truth distribution $\Prob( \cdot \given \bx)$ on the label space $\cY = \{ 0,1 \}^3$ are given as follows:

\begin{center}
\setlength\tabcolsep{3.5pt} 
\begin{tabular}{r|cccccccc}
\hline
$\by$ & $(0,0,0)$ & $(1,1,1)$ & $(0,1,1)$ & $(1,0,1)$ & $(1,1,0)$ \\
 $\Prob( \by \given \bx)$ & $\nicefrac{1}{4}$ & $\nicefrac{3}{16}$ & $\nicefrac{3}{16}$ & $\nicefrac{3}{16}$ & $\nicefrac{3}{16}$\\
 \hline
\end{tabular}
\end{center}

Obviously, the Bayes-optimal prediction for the subset 0/1 loss is $(0,0,0)$, and ideally, this prediction is produced by each classifier in the ensemble. Now, since these classifiers are not perfect, suppose that the different label combinations are predicted in proportion to their conditional probabilities, i.e., $(0,0,0)$ is predicted with probability $\nicefrac{1}{4}$, $(1,1,1)$ with probability $\nicefrac{3}{16}$, etc. One easily verifies that, for each of the three labels, the probability of it being predicted as relevant ($\nicefrac{9}{16}$) exceeds the probability for irrelevant ($\nicefrac{7}{16}$). Therefore, by taking the arithmetic average over the ensemble members' predictions, and then thresholding at $\nicefrac{1}{2}$, one will likely end up with the suboptimal prediction $(1,1,1)$.      

The reader may have noticed that this example is actually less problematic for the Hamming loss, for which the prediction $(1,1,1)$ is indeed Bayes-optimal, and would be produced by the label-wise aggregation sketched above. More generally, it is plausible that a label-wise combination of predictions is indeed suitable for decomposable losses like Hamming, but suboptimal for non-decomposable losses. 

Based on the discussion so far, we propose two aggregation functions for PTC, which can be seen as implementations of different types of voting, and will be used in our experimental study below:
\begin{itemize}
\item[•] \textbf{Label-wise voting (PTC-lw)}: For each individual label $\lambda_i$, the number of positive (relevant) and negative (irrelevant) votes in the predictions $\hat{\by}_1 , \ldots , \hat{\by}_M$ is counted, and the majority is adopted. 
\item[•] \textbf{Mode (PTC-mode)}: Counting is done at the level of the entire predictions, i.e., we predict the label combination $\hat{\by}$ that occurs most frequently:
\begin{equation}\label{eq:mode}
\hat{\by} = \operatorname*{argmax}_{\bar{\by}} \sum_{j=1}^M \,  \big\llbracket \hat{\by}_i = \bar{\by} \big\rrbracket  \, .
\end{equation}
In case the maximum is not unique, ties are broken by choosing the maximal prediction with the highest score 
\begin{equation}
s(\hat{\by}) = \sum_{k=1}^K \sum_{j=1}^M\big\llbracket \hat{y}_k = \hat{y}_{k,j} \big\rrbracket \, .
\end{equation}
\end{itemize}

\section{Experimental Evaluation}
\label{sec:exp}
We perform experiments on eight standard benchmark datasets (cf.\ Table~\ref{tab:datasets}) from the MULAN repository\footnote{\url{http://mulan.sourceforge.net/datasets.html}}, following a $10$-fold cross-validation procedure. Our primary goal is to confirm that the loss-based aggregation methods PTC and CTP outperform the commonly used voting techniques. Moreover, we conjecture that CTP performs better than PTC for decomposable losses, and PTC better than CTP for non-decomposable losses. This is because accurate marginal probabilities are of utmost importance for decomposable losses\,---\,which is exactly what CTP accomplishes through averaging label-wise predictions. Likewise, PTC is more apt at capturing label dependencies, which is important for non-decomposable losses, because it aggregates over several predictions tailored to the target loss (instead of producing only a single one, as CTP).

\begin{table}[t!]
\caption{Datasets used in the experiments}\label{tab:datasets}
\centering
\begin{tabular}{lcrrrr}
\hline
\# &Name    &  \# Inst. & \# Nom. Feat.& \# Num. Feat. & \# Lab. \\
\hline
1 &Cal500    &502    &0    &68   &174   \\
2 &Emotions  &593    &0    &72   &6  \\
3 &Scene     &2407   &0    &294  &6  \\ 
4 &Yeast     &2417   &0    &103  &14 \\
5 &Mediamill &43907  &0    &120  &101 \\
6 &Flags	 &194	 &9	   &10	 &7 \\
7 &Medical	 &978	 &1449 &0	 &45 \\
8 &Bibtex	 &7395	 &1836 &0	 &159 \\
\hline
\end{tabular}
\end{table}

We conducted three series of experiments using the ensemble methods EMODT, EBR, and ECC with their cardinality set to~$50$. We employed logistic regression as the base classifiers for EBR and ECC and let them produce probabilistic predictions. Thus, each ensemble member provides a complete marginal probability~vector. 

\begin{table}
\caption{Predictive performance (in percent) and rank (small number) of aggregation methods with respect to the Hamming loss, the subset 0/1 loss and the F1-measure.
}
\medskip
\label{tab:exp}
\centering
\setlength\tabcolsep{1.17pt}
\begin{tabular}{|l|l|r c r c r c r c r c r c r c r c |l|}
\cline{3-19}
\multicolumn{2}{c|}{}         
 & \multicolumn{2}{c}{\scriptsize Cal500}
 & \multicolumn{2}{c}{\scriptsize \shortstack{Emo-\\tions}}
 & \multicolumn{2}{c}{\scriptsize Scene}
 & \multicolumn{2}{c}{\scriptsize Yeast}
 & \multicolumn{2}{c}{\scriptsize Flags}
 & \multicolumn{2}{c}{\scriptsize \shortstack{Medi-\\cal}}
 & \multicolumn{2}{c}{\scriptsize Bibtex} 
 & \multicolumn{2}{c|}{\scriptsize \shortstack{Media-\\ mill}}
 & {\scriptsize \shortstack{Avg.\\ ranks}} \\
\cline{3-19}
\multicolumn{2}{c}{} & \multicolumn{17}{|c|}{\scriptsize \textbf{EMODT}}  \\
\cline{1-19}
\parbox[c]{3mm}{\multirow{5}{*}{\resizebox{!}{0.85\height}{\rotatebox[origin=c]{90}{\scriptsize Hamming loss $\downarrow$}}}}
& {\scriptsize GMV }
            & $\boldsymbol{13.65}$          & $\scriptstyle 1$ 
            & $18.72$                       & $\scriptstyle 2$ 
            & $9.45$                        & $\scriptstyle 3$ 
            & $19.57$                       & $\scriptstyle 2$ 
            & $\boldsymbol{23.48}$          & $\scriptstyle 1$ 
            & $1.55$                        & $\scriptstyle 2$ 
            & $1.27$                        & $\scriptstyle 3$ 
            & $2.67$                        & $\scriptstyle 3$ 
            & $2.13$ \\
& {\scriptsize BMV }
            & $13.75$                       & $\scriptstyle 2$ 
            & $\boldsymbol{18.52}$          & $\scriptstyle 1$ 
            & $9.11$                        & $\scriptstyle 2$ 
            & $\boldsymbol{19.38}$          & $\scriptstyle 1$ 
            & $25.09$                       & $\scriptstyle 3$ 
            & $1.59$                        & $\scriptstyle 3$ 
            & $\boldsymbol{1.25}$           & $\scriptstyle 1.5$ 
            & $\boldsymbol{2.65}$           & $\scriptstyle 1.5$ 
            & $\boldsymbol{1.88}$    \\
 & {\scriptsize CTP  }   & \multicolumn{17}{c|}{\scriptsize equivalent to GMV}  \\
 & {\scriptsize PTC-lw }  & \multicolumn{17}{c|}{\scriptsize equivalent to BMV}  \\
 & {\scriptsize PTC-mode }
            & $14.41$                       & $\scriptstyle 3$ 
            & $19.66$                       & $\scriptstyle 3$ 
            & $\boldsymbol{8.14}$           & $\scriptstyle 1$ 
            & $19.77$                       & $\scriptstyle 3$ 
            & $24.58$                       & $\scriptstyle 2$ 
            & $\boldsymbol{1.52}$           & $\scriptstyle 1$ 
            & $\boldsymbol{1.25}$           & $\scriptstyle 1.5$ 
            & $\boldsymbol{2.65}$           & $\scriptstyle 1.5$ 
            & $2.00$ \\
\cline{1-19}
\parbox[c]{3mm}{\multirow{5}{*}{\resizebox{!}{0.85\height}{\rotatebox[origin=c]{90}{\scriptsize Subset 0/1 loss $\downarrow$}}}}
& {\scriptsize GMV }
            & $\boldsymbol{100}$             & $\scriptstyle 2$ 
            & $69.82$                        & $\scriptstyle 3$ 
            & $49.48$                        & $\scriptstyle 3$ 
            & $85.48$                        & $\scriptstyle 3$ 
            & $79.34$                        & $\scriptstyle 2$ 
            & $55.70$                        & $\scriptstyle 2$ 
            & $87.59$                        & $\scriptstyle 3$ 
            & $84.79$                        & $\scriptstyle 3$ 
            & $2.63$ \\
& {\scriptsize BMV }
            & $\boldsymbol{100}$             & $\scriptstyle 2$ 
            & $67.81$                        & $\scriptstyle 2$ 
            & $46.16$                        & $\scriptstyle 2$
            & $83.62$                        & $\scriptstyle 2$ 
            & $83.00$                        & $\scriptstyle 3$ 
            & $57.26$                        & $\scriptstyle 3$ 
            & $86.61$                        & $\scriptstyle 2$ 
            & $84.57$                        & $\scriptstyle 2$ 
            & $2.25$ \\
& {\scriptsize CTP }    & \multicolumn{17}{c|}{\scriptsize equivalent to GMV} \\
& {\scriptsize PTC-lw } & \multicolumn{17}{c|}{\scriptsize equivalent to BMV} \\
& {\scriptsize PTC-mode }
            & $\boldsymbol{100}$             & $\scriptstyle 2$ 
            & $\boldsymbol{64.45}$           & $\scriptstyle 1$ 
            & $\boldsymbol{27.21}$           & $\scriptstyle 1$ 
            & $\boldsymbol{74.43}$           & $\scriptstyle 1$ 
            & $\boldsymbol{78.82}$           & $\scriptstyle 1$
            & $\boldsymbol{50.96}$           & $\scriptstyle 1$ 
            & $\boldsymbol{85.22}$           & $\scriptstyle 1$
            & $\boldsymbol{79.21}$           & $\scriptstyle 1$ 
            & $\boldsymbol{1.13}$ \\
\cline{1-19}
\parbox[c]{3mm}{\multirow{5}{*}{\resizebox{!}{0.85\height}{\rotatebox[origin=c]{90}{\scriptsize F1-measure $\uparrow$}}}}
& {\scriptsize GMV }
            & $33.31$                        & $\scriptstyle 5$ 
            & $58.18$                        & $\scriptstyle 5$
            & $53.45$                        & $\scriptstyle 5$ 
            & $58.53$                        & $\scriptstyle 5$ 
            & $74.91$                        & $\scriptstyle 4$ 
            & $51.17$                        & $\scriptstyle 4$ 
            & $25.33$                        & $\scriptstyle 5$ 
            & $59.52$                        & $\scriptstyle 5$ 
            & $4.75$ \\
&{\scriptsize BMV }
            & $37.33$                        & $\scriptstyle 4$ 
            & $61.79$                        & $\scriptstyle 4$ 
            & $57.02$                        & $\scriptstyle 4$ 
            & $60.99$                        & $\scriptstyle 4$ 
            & $74.33$                        & $\scriptstyle 5$ 
            & $50.72$                        & $\scriptstyle 5$ 
            & $27.69$                        & $\scriptstyle 4$ 
            & $60.74$                        & $\scriptstyle 4$ 
            & $4.25$ \\
&{\scriptsize CTP }
            & $\boldsymbol{48.31}$           & $\scriptstyle 1$ 
            & $\boldsymbol{68.30}$           & $\scriptstyle 1$ 
            & $74.90$                        & $\scriptstyle 2$ 
            & $\boldsymbol{66.61}$           & $\scriptstyle 1$ 
            & $75.89$                        & $\scriptstyle 3$ 
            & $\boldsymbol{76.17}$           & $\scriptstyle 1$ 
            & $\boldsymbol{48.77}$           & $\scriptstyle 1$ 
            & $\boldsymbol{63.88}$           & $\scriptstyle 1$ 
            & $\boldsymbol{1.38}$ \\
&{\scriptsize PTC-lw }
            & $46.45$                        & $\scriptstyle 2$
            & $68.29$                        & $\scriptstyle 2$ 
            & $71.78$                        & $\scriptstyle 3$ 
            & $64.87$                        & $\scriptstyle 3$ 
            & $76.21$                        & $\scriptstyle 2$ 
            & $71.81$                        & $\scriptstyle 3$ 
            & $37.79$                        & $\scriptstyle 3$
            & $62.71$                        & $\scriptstyle 2$ 
            & $2.50$ \\
&{\scriptsize PTC-mode}
            & $42.30$                        & $\scriptstyle 3$ 
            & $68.25$                        & $\scriptstyle 3$ 
            & $\boldsymbol{77.55}$           & $\scriptstyle 1$ 
            & $65.38$                        & $\scriptstyle 2$ 
            & $\boldsymbol{76.48}$           & $\scriptstyle 1$ 
            & $75.38$                        & $\scriptstyle 2$ 
            & $45.59$                        & $\scriptstyle 2$ 
            & $62.09$                        & $\scriptstyle 3$ 
            & $2.13$ \\
\cline{1-19}
\multicolumn{2}{c}{} & \multicolumn{17}{|c|}{\scriptsize \textbf{ECC}}  \\
\cline{1-19}
\parbox[c]{3mm}{\multirow{5}{*}{\resizebox{!}{0.85\height}{\rotatebox[origin=c]{90}{\scriptsize Hamming loss $\downarrow$}}}}
& {\scriptsize GMV }
            & $14.08$                        & $\scriptstyle 2$
            & $\boldsymbol{20.28}$           & $\scriptstyle 1$ 
            & $\boldsymbol{8.63}$            & $\scriptstyle 1$ 
            & $20.24$                        & $\scriptstyle 2$ 
            & $23.26$                        & $\scriptstyle 2$ 
            & $0.89$                         & $\scriptstyle 3$ 
            & $\boldsymbol{1.28}$            & $\scriptstyle 1$
            & \multicolumn{2}{c|}{---}
            & $\boldsymbol{1.71}$ \\
& {\scriptsize BMV }
            & $\boldsymbol{14.06}$           & $\scriptstyle 1$ 
            & $20.31$                        & $\scriptstyle 2$ 
            & $8.76$                         & $\scriptstyle 2$ 
            & $\boldsymbol{20.11}$           & $\scriptstyle 1$ 
            & $23.48$                        & $\scriptstyle 3$
            & $\boldsymbol{0.88}$            & $\scriptstyle 1.5$ 
            & $1.29$                         & $\scriptstyle 2.5$
            & \multicolumn{2}{c|}{---}
            & $1.86$ \\
 & {\scriptsize CTP  }   & \multicolumn{17}{c|}{\scriptsize equivalent to GMV}  \\
 & {\scriptsize PTC-lw }  & \multicolumn{17}{c|}{\scriptsize equivalent to BMV}  \\
 & {\scriptsize PTC-mode }
            & $14.24$                        & $\scriptstyle 3$ 
            & $20.48$                        & $\scriptstyle 3$ 
            & $8.77$                         & $\scriptstyle 3$ 
            & $20.39$                        & $\scriptstyle 3$ 
            & $\boldsymbol{22.52}$           & $\scriptstyle 1$ 
            & $\boldsymbol{0.88}$            & $\scriptstyle 1.5$ 
            & $1.29$                         & $\scriptstyle 2.5$
            & \multicolumn{2}{c|}{---}
            & $2.43$ \\
\cline{1-19}
\parbox[c]{3mm}{\multirow{5}{*}{\resizebox{!}{0.85\height}{\rotatebox[origin=c]{90}{\scriptsize Subset 0/1 loss $\downarrow$}}}}
& {\scriptsize GMV }
            & $\boldsymbol{100}$             & $\scriptstyle 2$ 
            & $69.46$                        & $\scriptstyle 2$ 
            & $32.82$                        & $\scriptstyle 2$ 
            & $79.35$                        & $\scriptstyle 3$ 
            & $72.32$                        & $\scriptstyle 2$ 
            & $29.49$                        & $\scriptstyle 3$ 
            & $81.61$                        & $\scriptstyle 2$
            & \multicolumn{2}{c|}{---}
            & $2.29$ \\
& {\scriptsize BMV }
            & $\boldsymbol{100}$             & $\scriptstyle 2$
            & $70.47$                        & $\scriptstyle 3$ 
            & $33.07$                        & $\scriptstyle 3$ 
            & $78.98$                        & $\scriptstyle 2$
            & $74.87$                        & $\scriptstyle 3$ 
            & $28.45$                        & $\scriptstyle 2$
            & $81.64$                        & $\scriptstyle 3$
            & \multicolumn{2}{c|}{---}
            & $2.57$ \\
& {\scriptsize CTP }    & \multicolumn{17}{c|}{\scriptsize equivalent to GMV} \\
& {\scriptsize PTC-lw } & \multicolumn{17}{c|}{\scriptsize equivalent to BMV} \\
& {\scriptsize PTC-mode }
            & $\boldsymbol{100}$             & $\scriptstyle 2$ 
            & $\boldsymbol{68.45}$           & $\scriptstyle 1$ 
            & $\boldsymbol{29.04}$           & $\scriptstyle 1$ 
            & $\boldsymbol{77.28}$           & $\scriptstyle 1$ 
            & $\boldsymbol{66.61}$           & $\scriptstyle 1$ 
            & $\boldsymbol{28.35}$           & $\scriptstyle 1$ 
            & $\boldsymbol{81.37}$           & $\scriptstyle 1$
            & \multicolumn{2}{c|}{---}
            & $\boldsymbol{1.14}$ \\
\cline{1-19}
\parbox[c]{3mm}{\multirow{5}{*}{\resizebox{!}{0.85\height}{\rotatebox[origin=c]{90}{\scriptsize F1-measure $\uparrow$}}}}
& {\scriptsize GMV }
            & $32.13$                        & $\scriptstyle 5$ 
            & $63.20$                        & $\scriptstyle 5$ 
            & $72.84$                        & $\scriptstyle 4$ 
            & $62.63$                        & $\scriptstyle 5$ 
            & $72.80$                        & $\scriptstyle 4$ 
            & $81.74$                        & $\scriptstyle 4$ 
            & $40.08$                        & $\scriptstyle 5$
            & \multicolumn{2}{c|}{---}
            & $4.57$ \\
&{\scriptsize BMV }
            & $32.50$                        & $\scriptstyle 4$ 
            & $64.00$                        & $\scriptstyle 4$ 
            & $71.75$                        & $\scriptstyle 5$ 
            & $62.81$                        & $\scriptstyle 4$ 
            & $72.56$                        & $\scriptstyle 5$ 
            & $81.10$                        & $\scriptstyle 5$ 
            & $40.10$                        & $\scriptstyle 4$
            & \multicolumn{2}{c|}{---}
            & $4.43$ \\
&{\scriptsize CTP }
            & $\boldsymbol{45.28}$           & $\scriptstyle 1$ 
            & $\boldsymbol{67.65}$           & $\scriptstyle 1$ 
            & $\boldsymbol{77.05}$           & $\scriptstyle 1$ 
            & $\boldsymbol{64.85}$           & $\scriptstyle 1$ 
            & $74.21$                        & $\scriptstyle 2$ 
            & $\boldsymbol{85.17}$           & $\scriptstyle 1$ 
            & $\boldsymbol{49.30}$           & $\scriptstyle 1$ 
            & \multicolumn{2}{c|}{---}
            & $\boldsymbol{1.14}$ \\
&{\scriptsize PTC-lw }
            & $42.05$                        & $\scriptstyle 2$ 
            & $67.58$                        & $\scriptstyle 2$
            & $75.52$                        & $\scriptstyle 3$ 
            & $64.69$                        & $\scriptstyle 2$ 
            & $\boldsymbol{74.53}$           & $\scriptstyle 1$ 
            & $84.85$                        & $\scriptstyle 2$ 
            & $48.95$                        & $\scriptstyle 2$
            & \multicolumn{2}{c|}{---}
            & $2.00$ \\
&{\scriptsize PTC-mode}
            & $41.98$                        & $\scriptstyle 3$ 
            & $66.81$                        & $\scriptstyle 3$
            & $75.57$                        & $\scriptstyle 2$
            & $64.06$                        & $\scriptstyle 3$ 
            & $74.13$                        & $\scriptstyle 3$ 
            & $84.58$                        & $\scriptstyle 3$ 
            & $48.77$                        & $\scriptstyle 3$
            & \multicolumn{2}{c|}{---}
            & $2.86$ \\
\cline{1-19}
\multicolumn{2}{c}{} & \multicolumn{17}{|c|}{\scriptsize \textbf{EBR}}  \\
\cline{1-19}
\parbox[c]{3mm}{\multirow{5}{*}{\resizebox{!}{0.85\height}{\rotatebox[origin=c]{90}{\scriptsize Hamming loss $\downarrow$}}}}
            & {\scriptsize GMV }
            & $\boldsymbol{13.95}$           & $\scriptstyle 1$ 
            & $\boldsymbol{20.15}$           & $\scriptstyle 1$ 
            & $9.82$                         & $\scriptstyle 2$ 
            & $19.89$                        & $\scriptstyle 3$ 
            & $24.05$                        & $\scriptstyle 2$ 
            & $\boldsymbol{0.92}$            & $\scriptstyle 2$ 
            & $\boldsymbol{1.22}$            & $\scriptstyle 1.5$
            & \multicolumn{2}{c|}{---}
            & $\boldsymbol{1.79}$ \\
& {\scriptsize BMV }
            & $14.02$                        & $\scriptstyle 2$ 
            & $20.32$                        & $\scriptstyle 3$ 
            & $9.83$                         & $\scriptstyle 3$ 
            & $\boldsymbol{19.87}$           & $\scriptstyle 1$ 
            & $\boldsymbol{23.62}$           & $\scriptstyle 1$ 
            & $\boldsymbol{0.92}$            & $\scriptstyle 2$
            & $\boldsymbol{1.22}$            & $\scriptstyle 1.5$
            & \multicolumn{2}{c|}{---}
            & $1.93$ \\
 & {\scriptsize CTP  }   & \multicolumn{17}{c|}{\scriptsize equivalent to GMV}  \\
 & {\scriptsize PTC-lw }  & \multicolumn{17}{c|}{\scriptsize equivalent to BMV}  \\
 & {\scriptsize PTC-mode }
            & $14.10$                        & $\scriptstyle 3$ 
            & $20.21$                        & $\scriptstyle 2$ 
            & $\boldsymbol{9.76}$            & $\scriptstyle 1$ 
            & $19.88$                        & $\scriptstyle 2$ 
            & $24.49$                        & $\scriptstyle 3$ 
            & $\boldsymbol{0.92}$            & $\scriptstyle 2$
            & $1.23$                         & $\scriptstyle 3$
            & \multicolumn{2}{c|}{---}
            & $2.29$ \\
\cline{1-19}
\parbox[c]{3mm}{\multirow{5}{*}{\resizebox{!}{0.85\height}{\rotatebox[origin=c]{90}{\scriptsize Subset 0/1 loss $\downarrow$}}}}
& {\scriptsize GMV }
            & $\boldsymbol{100}$             & $\scriptstyle 2$
            & $\boldsymbol{72.86}$           & $\scriptstyle 1$ 
            & $46.11$                        & $\scriptstyle 3$ 
            & $84.78$                        & $\scriptstyle 3$ 
            & $82.45$                        & $\scriptstyle 3$ 
            & $30.71$                        & $\scriptstyle 3$ 
            & $82.12$                        & $\scriptstyle 3$ 
            & \multicolumn{2}{c|}{---}
            & $2.57$ \\
& {\scriptsize BMV }
            & $\boldsymbol{100}$             & $\scriptstyle 2$ 
            & $73.53$                        & $\scriptstyle 3$ 
            & $46.03$                        & $\scriptstyle 2$ 
            & $84.49$                        & $\scriptstyle 2$ 
            & $81.92$                        & $\scriptstyle 2$ 
            & $30.39$                        & $\scriptstyle 2$ 
            & $81.88$                        & $\scriptstyle 2$ 
            & \multicolumn{2}{c|}{---}
            & $2.14$ \\
& {\scriptsize CTP }    & \multicolumn{17}{c|}{\scriptsize equivalent to GMV} \\
& {\scriptsize PTC-lw } & \multicolumn{17}{c|}{\scriptsize equivalent to BMV} \\
& {\scriptsize PTC-mode }
            & $\boldsymbol{100}$             & $\scriptstyle 2$ 
            & $73.37$                        & $\scriptstyle 2$
            & $\boldsymbol{45.53}$           & $\scriptstyle 1$ 
            & $\boldsymbol{84.32}$           & $\scriptstyle 1$ 
            & $\boldsymbol{80.87}$           & $\scriptstyle 1$ 
            & $\boldsymbol{30.09}$           & $\scriptstyle 1$ 
            & $\boldsymbol{81.72}$           & $\scriptstyle 1$ 
            & \multicolumn{2}{c|}{---}
            & $\boldsymbol{1.29}$ \\
\cline{1-19}
\parbox[c]{3mm}{\multirow{5}{*}{\resizebox{!}{0.85\height}{\rotatebox[origin=c]{90}{\scriptsize F1-measure $\uparrow$}}}}
& {\scriptsize GMV }
            & $34.17$                        & $\scriptstyle 5$ 
            & $58.38$                        & $\scriptstyle 4$ 
            & $61.99$                        & $\scriptstyle 5$ 
            & $61.37$                        & $\scriptstyle 5$ 
            & $72.60$                        & $\scriptstyle 4$ 
            & $79.18$                        & $\scriptstyle 5$ 
            & $39.75$                        & $\scriptstyle 5$
            & \multicolumn{2}{c|}{---}
            & $4.71$ \\
&{\scriptsize BMV }
            & $34.40$                        & $\scriptstyle 4$ 
            & $58.11$                        & $\scriptstyle 5$ 
            & $62.07$                        & $\scriptstyle 4$
            & $61.49$                        & $\scriptstyle 4$
            & $73.42$                        & $\scriptstyle 3$ 
            & $79.76$                        & $\scriptstyle 4$ 
            & $40.17$                        & $\scriptstyle 4$ 
            & \multicolumn{2}{c|}{---}
            & $4.00$ \\
&{\scriptsize CTP }
            & $\boldsymbol{47.62}$           & $\scriptstyle 1$
            & $66.67$                        & $\scriptstyle 3$ 
            & $76.14$                        & $\scriptstyle 3$ 
            & $\boldsymbol{65.07}$           & $\scriptstyle 1$ 
            & $\boldsymbol{75.18}$           & $\scriptstyle 1$ 
            & $\boldsymbol{85.27}$           & $\scriptstyle 1$ 
            & $\boldsymbol{50.78}$           & $\scriptstyle 1$ 
            & \multicolumn{2}{c|}{---}
            & $\boldsymbol{1.57}$ \\
&{\scriptsize PTC-lw }
            & $47.46$                        & $\scriptstyle 2$ 
            & $66.96$                        & $\scriptstyle 2$ 
            & $\boldsymbol{76.29}$           & $\scriptstyle 1$ 
            & $65.01$                        & $\scriptstyle 2$ 
            & $74.80$                        & $\scriptstyle 2$ 
            & $84.63$                        & $\scriptstyle 2$ 
            & $49.02$                        & $\scriptstyle 3$ 
            & \multicolumn{2}{c|}{---}
            & $2.00$ \\
&{\scriptsize PTC-mode}
            & $47.33$                        & $\scriptstyle 3$ 
            & $\boldsymbol{67.07}$           & $\scriptstyle 1$ 
            & $76.18$                        & $\scriptstyle 2$ 
            & $64.99$                        & $\scriptstyle 3$ 
            & $71.41$                        & $\scriptstyle 5$ 
            & $84.58$                        & $\scriptstyle 3$ 
            & $49.43$                        & $\scriptstyle 2$ 
            & \multicolumn{2}{c|}{---}
            & $2.71$ \\
\hline
\end{tabular}
\end{table}

The detailed  results are shown in Table \ref{tab:exp}. The best way for getting an insight into the respective performances is to consider the averages of these ranks in the final column of the table. In particular, each column shows the results of one dataset, each line shows the results of a combination of ensemble technique, loss function, and aggregation technique. For each combination of ensemble, loss function, and dataset, we also report the respective ranks for the obtained losses over the aggregation approaches. The bold value indicates the best performance on each data set. According to the Friedman/Nemenyi test, differences are statistically significant for a critical distance between the average ranks of 1.10/1.25 for $\alpha$=0.1/0.05 for EMODT, and similarly 1.94/2.16 and 2.08/2.31 for ECC and EBR, respectively. The Friedman test fails for all Hamming loss comparisons.

\begin{itemize}
\item[•] \textbf{Loss-based aggregation vs.\ voting.} Especially for F1 and subset 0/1 loss, there are (statistically significant) large differences between the voting-based decompositions on the one side, and PTC/CTP on the other side. This confirms our expectation that GMV and BMV are poorly suited for the case of non-decomposable performance measures. Only for Hamming loss, the voting-based techniques are in the same range, and, in fact, sometimes even better (yet, no significant difference). 

This result is also expected because Hamming loss is decomposable, so that its performance primarily depends on accurate marginal probabilities. In fact, in this case, label-wise PTC and  CTP are equivalent to binary and graded voting, respectively. 
For subset 0/1 loss, assuming label independence and marginal probability as the relevance information, our loss-based instantiations of CTP and PTC-lw are equivalent to GMV and BMV,~respectively. As can be seen from the results, however, this assumption is most likely invalid for the investigated datasets, because PTC-mode, which addresses the problem of finding the mode of the joint label distribution, typically outperforms the alternatives.
\item[•] \textbf{PTC vs.\ CTP.} With respect to the two different approaches, the mode-based PTC decomposition performs significantly better for subset 0/1 loss, whereas CTP (or, in this case, equivalently GMV) seems to perform better for Hamming loss. These results provide clear evidence in favor of our conjecture. The results for F1 are a bit more difficult to interpret but also consistent. 
Given marginal probabilities, we derive the loss minimizer for F1 under the assumption of label independence, and in this case, accurate marginal probabilities are again crucial. This is probably the reason for why CTP has an advantage over~PTC.
\end{itemize}

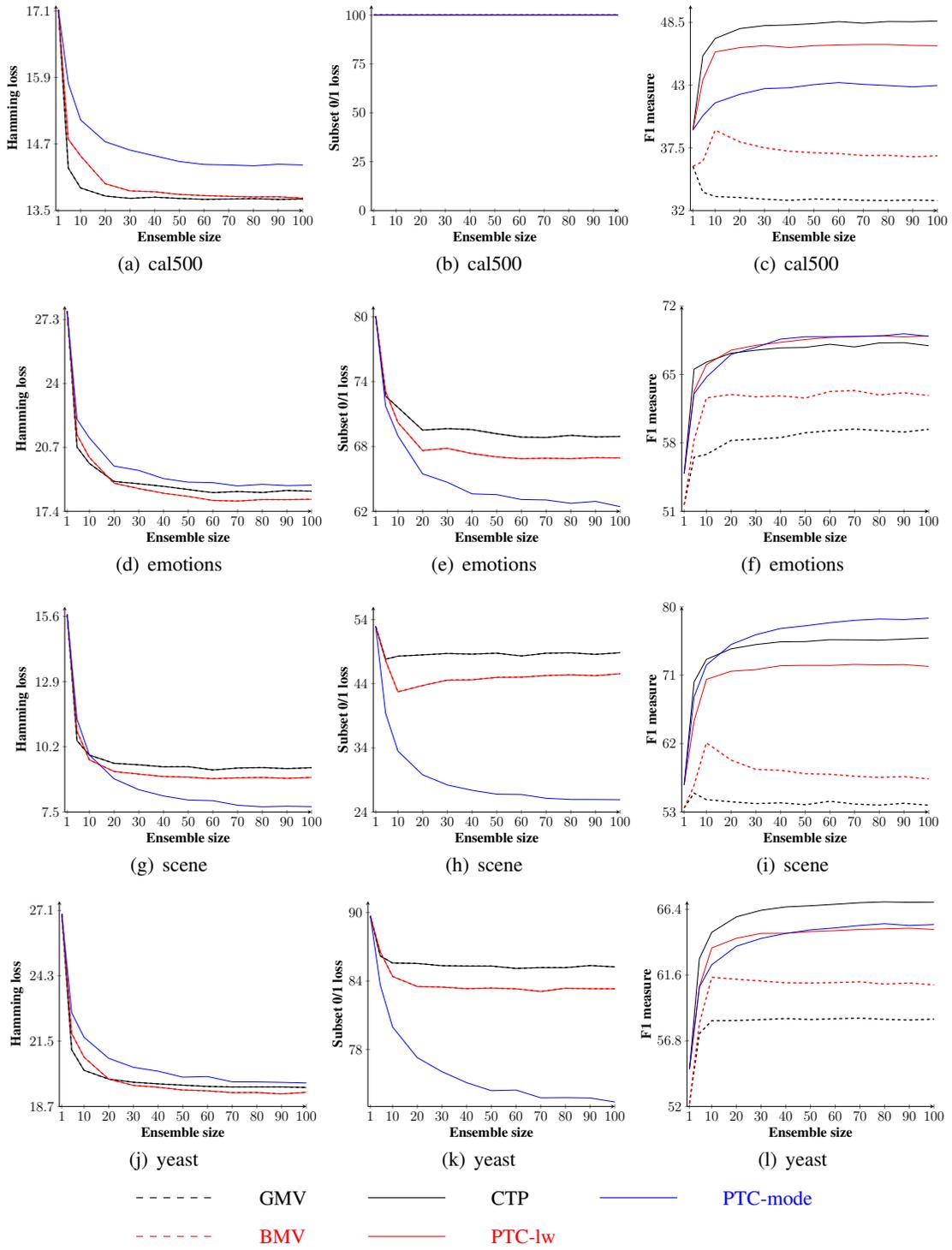
\begin{figure}
    \centering 
        \subfigure[cal500]{
        \begin{tikzpicture}[font=\Large, remember picture,scale=0.43]
            \begin{axis}[
                    xlabel={\textbf{Ensemble size}},
                    ylabel={\textbf{Hamming loss}},
                    axis lines=left,
                    xmin=0,
                    xmax=100,
                    ytick={13.5,14.7,...,17.2},
                    xtick={1,10,20,...,100},
                    ymin=13.5,
                    ymax=17.2,
                    thick,
                    width=0.7\textwidth,
                    height=0.6\textwidth
                ]
                \addplot[dashed] plot coordinates {  
                (1,	17.121	)(5,	14.268	)(10,	13.907	)(20,	13.762	)(30,	13.72	)(40,	13.742	)(50,	13.718	)(60,	13.702	)(70,	13.708	)(80,	13.714	)(90,	13.701	)(100,	13.708	)};                                                               
                \addplot[red, dashed] plot coordinates {
                (1,	17.121	)(5,	14.786	)(10,	14.485	)(20,	13.984	)(30,	13.854	)(40,	13.839	)(50,	13.79	)(60,	13.771	)(70,	13.757	)(80,	13.746	)(90,	13.749	)(100,	13.724	)};
                 \addplot[thick] plot coordinates {
                (1,	17.121	)(5,	14.268	)(10,	13.907	)(20,	13.762	)(30,	13.72	)(40,	13.742	)(50,	13.718	)(60,	13.702	)(70,	13.708	)(80,	13.714	)(90,	13.701	)(100,	13.708	)};
                 \addplot[red,thick] plot coordinates {
                 (1,	17.121	)(5,	14.786	)(10,	14.485	)(20,	13.984	)(30,	13.854	)(40,	13.839	)(50,	13.79	)(60,	13.771	)(70,	13.757	)(80,	13.746	)(90,	13.749	)(100,	13.724	)};
                 \addplot[blue,thick] plot coordinates {
                 (1,	17.121	)(5,	15.796	)(10,	15.134	)(20,	14.74	)(30,	14.59	)(40,	14.486	)(50,	14.384	)(60,	14.331	)(70,	14.322	)(80,	14.307	)(90,	14.335	)(100,	14.322	)};
            \end{axis}
        \end{tikzpicture}
    }%
        \subfigure[cal500]{
        \begin{tikzpicture}[font=\Large, remember picture,scale=0.43]
            \begin{axis}[
                    xlabel={\textbf{Ensemble size}},
                    ylabel={\textbf{Subset 0/1 loss}},
                    axis lines=left,
                    xmin=0,
                    xmax=100,
                    ytick={0,25,...,105},
                    xtick={1,10,20,...,100},
                    ymin=0,
                    ymax=105,
                    thick,
                    width=0.7\textwidth,
                    height=0.6\textwidth
                ]
                \addplot[dashed] plot coordinates {  
                (1,	100	)(5,	100	)(10,	100	)(20,	100	)(30,	100	)(40,	100	)(50,	100	)(60,	100	)(70,	100	)(80,	100	)(90,	100	)(100,	100	)};                                                               
                \addplot[red, dashed] plot coordinates {
                (1,	100	)(5,	100	)(10,	100	)(20,	100	)(30,	100	)(40,	100	)(50,	100	)(60,	100	)(70,	100	)(80,	100	)(90,	100	)(100,	100	)};
                 \addplot[thick] plot coordinates {
                (1,	100	)(5,	100	)(10,	100	)(20,	100	)(30,	100	)(40,	100	)(50,	100	)(60,	100	)(70,	100	)(80,	100	)(90,	100	)(100,	100	)};
                 \addplot[red,thick] plot coordinates {
                (1,	100	)(5,	100	)(10,	100	)(20,	100	)(30,	100	)(40,	100	)(50,	100	)(60,	100	)(70,	100	)(80,	100	)(90,	100	)(100,	100	)};
                 \addplot[blue,thick] plot coordinates {
                 (1,	100	)(5,	100	)(10,	100	)(20,	100	)(30,	100	)(40,	100	)(50,	100	)(60,	100	)(70,	100	)(80,	100	)(90,	100	)(100,	100	)};
            \end{axis}
        \end{tikzpicture}
    }%
        \subfigure[cal500]{
        \begin{tikzpicture}[font=\Large, remember picture,scale=0.43]
            \begin{axis}[
                    xlabel={\textbf{Ensemble size}},
                    ylabel={\textbf{F1 measure}},
                    axis lines=left,
                    xmin=0,
                    xmax=100,
                    ytick={32,37.5,...,50},
                    xtick={1,10,20,...,100},
                    ymin=32,
                    ymax=50,
                    thick,
                    width=0.7\textwidth,
                    height=0.6\textwidth
                ]
                \addplot[dashed] plot coordinates {  
                (1,	35.878	)(5,	33.62	)(10,	33.236	)(20,	33.142	)(30,	33	)(40,	32.895	)(50,	33.017	)(60,	32.975	)(70,	32.898	)(80,	32.88	)(90,	32.924	)(100,	32.877	)};                                                               
                \addplot[red, dashed] plot coordinates {
                (1,	35.878	)(5,	36.313	)(10,	39.047	)(20,	38.019	)(30,	37.504	)(40,	37.205	)(50,	37.07	)(60,	37.002	)(70,	36.823	)(80,	36.846	)(90,	36.718	)(100,	36.801	)};
                 \addplot[thick] plot coordinates {
                (1,	39.071	)(5,	45.571	)(10,	47.088	)(20,	47.956	)(30,	48.221	)(40,	48.277	)(50,	48.395	)(60,	48.57	)(70,	48.43	)(80,	48.576	)(90,	48.559	)(100,	48.625	)};
                 \addplot[red,thick] plot coordinates {
                (1,	39.071	)(5,	43.468	)(10,	45.911	)(20,	46.293	)(30,	46.468	)(40,	46.297	)(50,	46.46	)(60,	46.522	)(70,	46.564	)(80,	46.564	)(90,	46.48	)(100,	46.454	)};
                 \addplot[blue,thick] plot coordinates {
                (1,	39.071	)(5,	40.317	)(10,	41.447	)(20,	42.2	)(30,	42.697	)(40,	42.772	)(50,	43.05	)(60,	43.227	)(70,	43.07	)(80,	42.968	)(90,	42.851	)(100,	42.974	)};
            \end{axis}
        \end{tikzpicture}
    }%
 \qquad   
        \subfigure[emotions]{
        \begin{tikzpicture}[font=\Large, remember picture,scale=0.43]
        \begin{axis}[
                    xlabel={\textbf{Ensemble size}},
                    ylabel={\textbf{Hamming loss}},
                    axis lines=left,
                    xmin=0,
                    xmax=100,
                    ytick={17.4,20.7,...,28},
                    xtick={1,10,20,...,100},
                    ymin=17.4,
                    ymax=28,
                    thick,
                    width=0.7\textwidth,
                    height=0.6\textwidth
                ]
                \addplot[dashed] plot coordinates {  
                (1,27.742)	(5,20.714)	(10,19.87)	(20,18.94)	(30,18.819)	(40,18.688)	(50,18.522)	(60,18.363)	(70,18.419)	(80,18.369)	(90,18.475)	(100,18.436)};                                                               
                \addplot[red, dashed] plot coordinates {
                (1, 27.742)	(5,21.349)	(10,20.178)	(20,18.849)	(30,18.572)	(40,18.334)	(50,18.166)	(60,17.962)	(70,17.928)	(80,18.006)	(90,18)	(100,18.02)};
                 \addplot[thick] plot coordinates {
                (1,27.742)	(5,20.714)	(10,19.87)	(20,18.94)	(30,18.819)	(40,18.688)	(50,18.522)	(60,18.363)	(70,18.419)	(80,18.369)	(90,18.475)	(100,18.436)};
                 \addplot[red,thick] plot coordinates {
                 (1, 27.742)	(5,21.349)	(10,20.178)	(20,18.849)	(30,18.572)	(40,18.334)	(50,18.166)	(60,17.962)	(70,17.928)	(80,18.006)	(90,18)	(100,18.02)};
                 \addplot[blue,thick] plot coordinates {
                 (1, 27.742)	(5,22.151)	(10,21.193)	(20,19.734)	(30,19.51)	(40,19.088)	(50,18.903)	(60,18.877)	(70,18.709)	(80,18.791)	(90,18.723)	(100,18.749)};
            \end{axis}
        \end{tikzpicture}
    }%
        \subfigure[emotions]{
        \begin{tikzpicture}[font=\Large, remember picture,scale=0.43]
            \begin{axis}[
                    xlabel={\textbf{Ensemble size}},
                    ylabel={\textbf{Subset 0/1 loss}},
                    axis lines=left,
                    xmin=0,
                    xmax=100,
                    ytick={62,68,...,81},
                    xtick={1,10,20,...,100},
                    ymin=62,
                    ymax=81,
                    thick,
                    width=0.7\textwidth,
                    height=0.6\textwidth
                ]
                \addplot[dashed] plot coordinates {  
                (1,80.065)	(5,72.6629)	(10,71.618)	(20,69.513)	(30,69.65)	(40,69.563)	(50,69.173)	(60,68.869)	(70,68.831)	(80,69.028)	(90,68.886)	(100,68.925)};                                                               
                \addplot[red, dashed] plot coordinates {
                (1, 80.065)	(5,73.098)	(10,70.22)	(20,67.626)	(30,67.826)	(40,67.342)	(50,67.0319)	(60,66.867)	(70,66.91)	(80,66.868)	(90,66.966)	(100,66.934)};
                 \addplot[thick] plot coordinates {
                (1,80.065)	(5,72.6629)	(10,71.618)	(20,69.513)	(30,69.65)	(40,69.563)	(50,69.173)	(60,68.869)	(70,68.831)	(80,69.028)	(90,68.886)	(100,68.925)};
                 \addplot[red,thick] plot coordinates {
                (1, 80.065)	(5,73.098)	(10,70.22)	(20,67.626)	(30,67.826)	(40,67.342)	(50,67.0319)	(60,66.867)	(70,66.91)	(80,66.868)	(90,66.966)	(100,66.934)};
                 \addplot[blue,thick] plot coordinates {
                 (1, 80.065)	(5,71.738)	(10,68.986)	(20,65.463)	(30,64.684)	(40,63.613)	(50,63.54)	(60,63.09)	(70,63.048)	(80,62.736)	(90,62.917)	(100,62.431)};
            \end{axis}
        \end{tikzpicture}
    }%
        \subfigure[emotions]{
        \begin{tikzpicture}[font=\Large, remember picture,scale=0.43]
            \begin{axis}[
                    xlabel={\textbf{Ensemble size}},
                    ylabel={\textbf{F1 measure}},
                    axis lines=left,
                    xmin=0,
                    xmax=100,
                    ytick={51,58,...,72},
                    xtick={1,10,20,...,100},
                    ymin=51,
                    ymax=72,
                    thick,
                    width=0.7\textwidth,
                    height=0.6\textwidth
                ]
                \addplot[dashed] plot coordinates {  
                (1,51.678)	(5,56.525)	(10,56.813)	(20,58.248)	(30,58.383)	(40,58.54)	(50,59.028)	(60,59.244)	(70,59.412)	(80,59.262)	(90,59.107)	(100,59.384)};                                                               
                \addplot[red, dashed] plot coordinates {
                (1, 51.678)	(5,58.279)	(10,62.605)	(20,62.955)	(30,62.713)	(40,62.813)	(50,62.587)	(60,63.257)	(70,63.363)	(80,62.9)	(90,63.128)	(100,62.843)};
                 \addplot[thick] plot coordinates {
                (1,54.852)	(5,65.535)	(10,66.257)	(20,67.171)	(30,67.477)	(40,67.714)	(50,67.761)	(60,68.088)	(70,67.81)	(80,68.223)	(90,68.247)	(100,67.941)};
                 \addplot[red,thick] plot coordinates {
                (1, 54.852)	(5,63.3)	(10,66.002)	(20,67.482)	(30,67.976)	(40,68.279)	(50,68.572)	(60,68.787)	(70,68.882)	(80,68.964)	(90,68.83)	(100,68.947)};
                 \addplot[blue,thick] plot coordinates {
                 (1, 54.852)	(5,63.023)	(10,64.716)	(20,67.049)	(30,67.767)	(40,68.611)	(50,68.852)	(60,68.857)	(70,68.874)	(80,68.921)	(90,69.159)	(100,68.895)};
            \end{axis}
        \end{tikzpicture}
    }%
 \qquad
         \subfigure[scene]{
        \begin{tikzpicture}[font=\Large, remember picture,scale=0.43]
            \begin{axis}[
                    xlabel={\textbf{Ensemble size}},
                    ylabel={\textbf{Hamming loss}},
                    axis lines=left,
                    xmin=0,
                    xmax=100,
                    ytick={7.5,10.2,...,16},
                    xtick={1,10,20,...,100},
                    ymin=7.5,
                    ymax=16,
                    thick,
                    width=0.7\textwidth,
                    height=0.6\textwidth
                ]
                \addplot[dashed] plot coordinates {  
                (1,15.673)	(5,10.452)	(10,9.854)	(20,9.515)	(30,9.454)	(40,9.37)	(50,9.375)	(60,9.24)	(70,9.32)	(80,9.339)	(90,9.295)	(100,9.334)};                                                               
                \addplot[red, dashed] plot coordinates {
                (1, 15.673)	(5,10.845)	(10,9.656)	(20,9.175)	(30,9.068)	(40,8.968)	(50,8.941)	(60,8.878)	(70,8.913)	(80,8.929)	(90,8.893)	(100,8.93)};
                 \addplot[thick] plot coordinates {
                (1,15.673)	(5,10.452)	(10,9.854)	(20,9.515)	(30,9.454)	(40,9.37)	(50,9.375)	(60,9.24)	(70,9.32)	(80,9.339)	(90,9.295)	(100,9.334)};  
                 \addplot[red,thick] plot coordinates {
                (1, 15.673)	(5,10.845)	(10,9.656)	(20,9.175)	(30,9.068)	(40,8.968)	(50,8.941)	(60,8.878)	(70,8.913)	(80,8.929)	(90,8.893)	(100,8.93)};
                 \addplot[blue,thick] plot coordinates {
                 (1, 15.673)	(5,11.345)	(10,9.86)	(20,8.873)	(30,8.423)	(40,8.163)	(50,7.996)	(60,7.968)	(70,7.78)	(80,7.715)	(90,7.74)	(100,7.72)};
            \end{axis}
        \end{tikzpicture}
    }%
        \subfigure[ scene]{
        \begin{tikzpicture}[font=\Large, remember picture,scale=0.43]
        \begin{axis}[
                    xlabel={\textbf{Ensemble size}},
                    ylabel={\textbf{Subset 0/1 loss}},
                    axis lines=left,
                    xmin=0,
                    xmax=100,
                    ytick={24,34,...,56},
                    xtick={1,10,20,...,100},
                    ymin=24,
                    ymax=56,
                    thick,
                    width=0.7\textwidth,
                    height=0.6\textwidth
                ]
                \addplot[dashed] plot coordinates {  
                (1,52.921)	(5,47.814)	(10,48.292)	(20,48.493)	(30,48.709)	(40,48.608)	(50,48.759)	(60,48.319)	(70,48.755)	(80,48.815)	(90,48.575)	(100,48.831)};                                                               
                \addplot[red, dashed] plot coordinates {
                (1, 52.921)	(5,47.599)	(10,42.738)	(20,43.72)	(30,44.571)	(40,44.613)	(50,44.989)	(60,45.028)	(70,45.28)	(80,45.392)	(90,45.263)	(100,45.557)};
                 \addplot[thick] plot coordinates {
                (1,52.921)	(5,47.814)	(10,48.292)	(20,48.493)	(30,48.709)	(40,48.608)	(50,48.759)	(60,48.319)	(70,48.755)	(80,48.815)	(90,48.575)	(100,48.831)};                
                 \addplot[red,thick] plot coordinates {
                (1, 52.921)	(5,47.599)	(10,42.738)	(20,43.72)	(30,44.571)	(40,44.613)	(50,44.989)	(60,45.028)	(70,45.28)	(80,45.392)	(90,45.263)	(100,45.557)};
                 \addplot[blue,thick] plot coordinates {
                 (1, 52.921)	(5,39.397)	(10,33.501)	(20,29.793)	(30,28.213)	(40,27.383)	(50,26.773)	(60,26.724)	(70,26.141)	(80,25.949)	(90,25.945)	(100,25.912)};
            \end{axis}
        \end{tikzpicture}
    }%
        \subfigure[ scene]{
        \begin{tikzpicture}[font=\Large, remember picture,scale=0.43]
        \begin{axis}[
                    xlabel={\textbf{Ensemble size}},
                    ylabel={\textbf{F1 measure}},
                    axis lines=left,
                    xmin=0,
                    xmax=100,
                    ytick={53,62,...,80},
                    xtick={1,10,20,...,100},
                    ymin=53,
                    ymax=80,
                    thick,
                    width=0.7\textwidth,
                    height=0.6\textwidth
                ]
                \addplot[dashed] plot coordinates {  
                (1,53.527)	(5,55.48)	(10,54.624)	(20,54.348)	(30,54.101)	(40,54.205)	(50,53.955)	(60,54.413)	(70,54.046)	(80,53.904)	(90,54.146)	(100,53.89)};                                                               
                \addplot[red, dashed] plot coordinates {
                (1, 53.527)	(5,56.495)	(10,62.107)	(20,59.823)	(30,58.607)	(40,58.481)	(50,58.033)	(60,57.958)	(70,57.703)	(80,57.541)	(90,57.647)	(100,57.349)};
                 \addplot[thick] plot coordinates {
                (1,56.567)	(5,70.1)	(10,73.073)	(20,74.457)	(30,75.018)	(40,75.382)	(50,75.411)	(60,75.659)	(70,75.638)	(80,75.599)	(90,75.739)	(100,75.88)};
                 \addplot[red,thick] plot coordinates {
                (1, 56.567)	(5,64.943)	(10,70.445)	(20,71.516)	(30,71.738)	(40,72.244)	(50,72.291)	(60,72.278)	(70,72.418)	(80,72.356)	(90,72.391)	(100,72.151)};
                 \addplot[blue,thick] plot coordinates {
                 (1, 56.567)	(5,68.143)	(10,72.33)	(20,75.027)	(30,76.308)	(40,77.132)	(50,77.484)	(60,77.891)	(70,78.208)	(80,78.381)	(90,78.31)	(100,78.495)};
            \end{axis}
        \end{tikzpicture}
    }%
    \qquad
         \subfigure[ yeast]{
        \begin{tikzpicture}[font=\Large, remember picture,scale=0.43]
        \begin{axis}[
                    xlabel={\textbf{Ensemble size}},
                    ylabel={\textbf{Hamming loss}},
                    axis lines=left,
                    xmin=0,
                    xmax=100,
                    ytick={18.7,21.5,...,27.5},
                    xtick={1,10,20,...,100},
                    ymin=18.7,
                    ymax=27.5,
                    thick,
                    width=0.7\textwidth,
                    height=0.6\textwidth
                ]
                \addplot[dashed] plot coordinates {  
                (1,26.944)	(5,21.127)	(10,20.249)	(20,19.87)	(30,19.737)	(40,19.667)	(50,19.612)	(60,19.557)	(70,19.534)	(80,19.536)	(90,19.535)	(100,19.514)};                                                               
                \addplot[red, dashed] plot coordinates {
                (1, 26.944)	(5,21.826)	(10,20.814)	(20,19.868)	(30,19.602)	(40,19.522)	(50,19.405)	(60,19.368)	(70,19.293)	(80,19.298)	(90,19.236)	(100,19.306)};
                 \addplot[thick] plot coordinates {
                (1,26.944)	(5,21.127)	(10,20.249)	(20,19.87)	(30,19.737)	(40,19.667)	(50,19.612)	(60,19.557)	(70,19.534)	(80,19.536)	(90,19.535)	(100,19.514)};   
                 \addplot[red,thick] plot coordinates {
                (1, 26.944)	(5,21.826)	(10,20.814)	(20,19.868)	(30,19.602)	(40,19.522)	(50,19.405)	(60,19.368)	(70,19.293)	(80,19.298)	(90,19.236)	(100,19.306)};
                 \addplot[blue,thick] plot coordinates {
                 (1, 26.944) (5,22.719)	(10,21.665)	(20,20.761)	(30,20.376)	(40,20.21)	(50,19.954)	(60,19.98)	(70,19.754)	(80,19.751)	(90,19.732)	(100,19.706)};
            \end{axis}
        \end{tikzpicture}
    }%
        \subfigure[ yeast]{
        \begin{tikzpicture}[font=\Large, remember picture,scale=0.43]
        \begin{axis}[
                    xlabel={\textbf{Ensemble size}},
                    ylabel={\textbf{Subset 0/1 loss}},
                    axis lines=left,
                    xmin=0,
                    xmax=100,
                    ytick={72,78,...,91},
                    xtick={1,10,20,...,100},
                    ymin=73,
                    ymax=91,
                    thick,
                    width=0.7\textwidth,
                    height=0.6\textwidth
                ]
                \addplot[dashed] plot coordinates {  
                (1,89.694)	(5,86.197)	(10,85.58)	(20,85.544)	(30,85.342)	(40,85.313)	(50,85.312)	(60,85.106)	(70,85.176)	(80,85.172)	(90,85.357)	(100,85.239)};                                                               
                \addplot[red, dashed] plot coordinates {
                (1, 89.694)	(5,86.551)	(10,84.401)	(20,83.523)	(30,83.472)	(40,83.331)	(50,83.386)	(60,83.317)	(70,83.09)	(80,83.369)	(90,83.328)	(100,83.319)};
                 \addplot[thick] plot coordinates {
                 (1,89.694)	(5,86.197)	(10,85.58)	(20,85.544)	(30,85.342)	(40,85.313)	(50,85.312)	(60,85.106)	(70,85.176)	(80,85.172)	(90,85.357)	(100,85.239)};               
                 \addplot[red,thick] plot coordinates {
                (1, 89.694)	(5,86.551)	(10,84.401)	(20,83.523)	(30,83.472)	(40,83.331)	(50,83.386)	(60,83.317)	(70,83.09)	(80,83.369)	(90,83.328)	(100,83.319)};
                 \addplot[blue,thick] plot coordinates {
                 (1, 89.694) (5,83.653)	(10,79.969)	(20,77.277)	(30,76.048)	(40,75.081)	(50,74.383)	(60,74.436)	(70,73.758)	(80,73.774)	(90,73.743)	(100,73.393)};                
            \end{axis}
        \end{tikzpicture}
    }%
        \subfigure[ yeast]{
        \begin{tikzpicture}[font=\Large, remember picture,scale=0.43]
        \begin{axis}[
                    xlabel={\textbf{Ensemble size}},
                    ylabel={\textbf{F1 measure}},
                    axis lines=left,
                    xmin=0,
                    xmax=100,
                    ytick={52,56.8,...,67},
                    xtick={1,10,20,...,100},
                    ymin=52,
                    ymax=67,
                    thick,
                    width=0.7\textwidth,
                    height=0.6\textwidth
                ]
                \addplot[dashed] plot coordinates {  
                (1,52.223)	(5,57.32)	(10,58.264)	(20,58.282)	(30,58.344)	(40,58.425)	(50,58.353)	(60,58.412)	(70,58.453)	(80,58.364)	(90,58.319)	(100,58.38)};                                                               
                \addplot[red, dashed] plot coordinates {
                (1, 52.223)	(5,58.046)	(10,61.437)	(20,61.298)	(30,61.172)	(40,61.037)	(50,61.022)	(60,61.052)	(70,61.111)	(80,60.937)	(90,61.029)	(100,60.879)};
                 \addplot[thick] plot coordinates {
                (1,54.737)	(5,62.803)	(10,64.722)	(20,65.853)	(30,66.335)	(40,66.579)	(50,66.664)	(60,66.773)	(70,66.888)	(80,66.948)	(90,66.916)	(100,66.931)};
                 \addplot[red,thick] plot coordinates {
                (1, 54.737)	(5,60.753)	(10,63.581)	(20,64.285)	(30,64.648)	(40,64.653)	(50,64.776)	(60,64.841)	(70,64.933)	(80,64.975)	(90,65.027)	(100,64.928)};
                 \addplot[blue,thick] plot coordinates {
                (1, 54.737)	(5,60.732)	(10,62.353)	(20,63.706)	(30,64.276)	(40,64.642)	(50,64.901)	(60,65.047)	(70,65.225)	(80,65.349)	(90,65.222)	(100,65.289)};
            \end{axis}
        \end{tikzpicture}
    }%
 \qquad
        \begin{tikzpicture}[remember picture,scale=0.6]
            \draw[red, dashed] (1,1)--(3,1);
            \node[right,red] at (4,1) { BMV};
            \draw[red] (7,1)--(9,1);
            \node[right,red] at (10,1) { PTC-lw};
            \draw[blue] (13,2)--(15,2);
            \node[right, blue] at (16,2) { PTC-mode};
            \draw[black, dashed] (1,2)--(3,2);
            \node[right] at (4,2) { GMV};
            \draw[black] (7,2)--(9,2);
            \node[right] at (10,2) { CTP};
             \end{tikzpicture}
           \caption{Predictive performance ($y$-axis) of aggregation methods as a function of the cardinality of ensembles ($x$-axis) in terms of Hamming loss (left column), subset 0/1 loss (middle), and F1 (right column) for four datasets.}
          \label{fig:EMLC_EMODT}
\end{figure}

We also conducted a series of experiments using EMODT 
with the number of ensemble members varying from $1$ to $100$ ($M \in \{1, 5, 10, 20, 30, \dots,100\}$). Here, our interest was to study the influence of the ensemble size on the performance of the aggregation methods. For each value of the ensemble cardinality, we have run a $10$ times $10$-fold cross-validation, for which we report the average scores. 
As expected, the results shown in Figure~\ref{fig:EMLC_EMODT} confirm that the MLC scores typically improve with an increasing size of the ensembles. This is in agreement with the observation on the performance of ECC reported in \cite{li2013selective}. More importantly, we also see differences between the different aggregation methods, and that suitable instantiations of CTP and PTC can indeed reach better performance than standard voting techniques. 
In particular, the visible gaps for the subset 0/1 loss re-confirm the superiority of PTC-mode for non-decomposable losses.
Finally, we note that the performances change rapidly in the beginning and tend to converge when the number of ensemble members reaches moderate values (i.e., $30$ or $40$), except for the subset 0/1 loss and PTC-mode. This is again in agreement with our expectations, because PTC-mode does voting at the level of the entire predictions, and the number of possible predictions increases exponentially with the number of labels, so that more iterations are necessary for convergence. 
A similar effect can be observed for PTC-lw/BMV, whose label-wise votings converge less rapidly to accurate marginal probability estimates than CTP/GMV, but are able to catch up with increasing number of votes. 
For EMODT, there seems to be even an advantage in the end for using the vote distributions, possibly due to less accurate probability estimates of the trees.
Results similar to those shown in Figure~\ref{fig:EMLC_EMODT} have been obtained for EBR and ECC.


\section{Conclusion}
\label{sec:con}

This paper studied the question of how to aggregate the predictions of individual members of an ensemble of multilabel classifiers in a systematic way. We introduced a formal framework of ensemble multi-label classification, in which we distinguish two principal approaches, referred to as ``predict then combine'' (PTC) and ``combine then predict'' (CTP). Both approaches generalize voting techniques commonly used for EMLC, while allowing one to explicitly take the target performance measure into account. Our framework supports the analysis of existing EMLC methods as well as the systematic development of new ones. Besides, it suggests a number of interesting theoretical problems, like the question of how to combine predictions in PTC in a provably optimal way. Experimentally, we showed that standard voting techniques are indeed outperformed by suitable instantiations of CTP and PTC. Moreover, our results suggest that CTP performs well for decomposable loss functions, whereas PTC is the better choice for non-decomposable losses.  



\end{document}